# PROBABILISTIC BLOCK TERM DECOMPOSITION FOR THE MODELING OF HIGHER-ORDER ARRAYS

**Jesper Løve Hinrich, and Morten Mørup**†

October 4, 2023

## ABSTRACT

Tensors are ubiquitous in science and engineering and tensor factorization approaches have become important tools for the characterization of higher order structure. Factorizations includes the outer-product rank Canonical Polyadic Decomposition (CPD) as well as the multi-linear rank Tucker decomposition in which the Block-Term Decomposition (BTD) is a structured intermediate interpolating between these two representations. Whereas CPD, Tucker, and BTD have traditionally relied on maximum-likelihood estimation, Bayesian inference has been use to form probabilistic CPD and Tucker. We propose, an efficient variational Bayesian probabilistic BTD, which uses the von-Mises Fisher matrix distribution to impose orthogonality in the multi-linear Tucker parts forming the BTD. On synthetic and two real datasets, we highlight the Bayesian inference procedure and demonstrate using the proposed pBTD on noisy data and for model order quantification. We find that the probabilistic BTD can quantify suitable multi-linear structures providing a means for robust inference of patterns in multi-linear data.

Bayesian inference, probabilistic, decomposition, factorization, tensor.

## 1 Introduction

Tensors or multi-way arrays naturally occur in practically all areas of science including psychology (i.e., human responses to questionnaire data according to scoring criteria of different objects), chemometrics (i.e., excitation and emission spectra across samples), biology (i.e., genetic expression of cell profiles across time and experimental conditions), and knowledge representations (i.e., entity-entity relationships across predicates), see also [1] and references therein. To analyze these multi-way arrays accounting for their higher order structure tensor decompositions have become important tools to characterize and discover structure in these data, see [2, 1] for details.

Tensor decompositions have historically focused on maximum likelihood estimation methods to obtain a point estimate to decompose the data, most predominately based on Gaussian likelihood (least squares estimation). Recently, there has been a rise in the development of Bayesian inference for tensor data, initially focusing on binary or count data, but now applied more broadly to various types of data, for an overview see [3, 4]. The benefits of a Bayesian approach are that it characterizes the decomposition solution as a distribution, the so-called posterior distribution, which allows characterization of the uncertainty whereas priors acts as regularizers adding robustness and preventing issues of degeneracy. Additionally, it provides a principled way to incorporate a priori information. For a review on maximum likelihood based and Bayesian tensor decomposition, see [2] and [3], respectively.

The two most common tensor decomposition methods are the Canonical Polyadic Decomposition/PARAFAC (CPD) and Tucker model. The CPD model represents the data through a sum of outer product rank-1 terms (i.e., separate multi-linear structures), whereas Tucker uses a multi-linear rank decomposition (i.e., with "connected" multi-linear structures). Theoretically, the CPD model can be considered a special case of the Tucker model in which the core-array accounting for interaction among factors of the different modes of the tensor is constrained to be (hyper-) diagonal. CPD is advantages in that its global optimum is unique up to scaling and permutation under mild conditions which is not the case for the unconstrained Tucker model that suffers from rotational indeterminacies, see also [2, 1]. The block-term decomposition (BTD) [5] can be considered a modeling framework specifying intermediate representations to the CPD and Tucker defined by aggregating multiple small Tucker models with specific multi-linear rank properties such that uniqueness can be guaranteed across blocks [6].

---

*J. L. Hinrich was with the Department of Food Science, University of Copenhagen, Denmark.

†M. Mørup was with the Department of Applied Mathematics and Computer Science, Technical University of Denmark, Denmark.

A key challenge in BTD is determining the number of cores and their sizes. Usually, the cores are set to be the same size which simplifies tuning this complexity parameter whereas exhaustive evaluation is computationally demanding. Unfortunately, there are in general only heuristic criteria for choosing between the different parameter-setting, see also [7] and references therein. The Bayesian framework here provides access to principled tools for regularization and order determination through prior specification (denoted automatic relevance determination (ARD)) and estimates of the marginal likelihood (model evidence) which has been shown to add robustness to model specification and noise for CPD, Tucker, and Tensor Train decomposition, see also [3] and references therein.

Recently, [8] proposed a Bayesian Block-Term Decomposition with Gaussian factor distributions, but their model does not inherit important inferential properties of BTD, as it ignores the orthogonality property central to the maximum likelihood model with the associated decoupling of core-elements. Additionally, their focus was on a special type of BTD - a so-called $(L_r, L_r, 1)$ BTD model. Here, we consider a general multi-linear rank Bayesian BTD with factors following a von Mises-Fisher distribution (orthogonal matrices) and use automatic relevance determination on the core array. We describe the model and inferential properties, then test it on both synthetically generated BTD data and two real data sets highlighting the use of Bayesian Tensor Factorization in the context of BTD including CPD and Tucker as limiting cases in which each block term respectively is rank-1 and only one block-term imposed of full multi-linear rank.

In summary, we provide a unified Bayesian framework for CPD, Tucker and BTD through the proposed probabilistic Block Term Decomposition (pBTD). We further highlight the use of Bayesian inference for tensor decomposition in the context of these prominent models in terms of model order assessment, robustness to noise, and efficient inference with complexity similar to the corresponding conventional maximum-likelihood estimation procedures. In particular, this paper highlights Bayesian inference for tensor decomposition corroborating the many existing efforts to endow tensor models uncertainty quantification presently expanding these efforts to efficiently and principled encompass the BTD.

## 2 Methods

In the following, scalars are indicated by lowercase letters $x$, vectors by bold lowercase letters $\mathbf{x}^{I \times 1}$, matrices by bold uppercase letters $\mathbf{X}^{I \times J}$, and $N^{\text{th}}$ order tensors by bold calligraphic letters $\boldsymbol{\mathcal{X}}^{I_1 \times I_2 \times \cdots I_N}$. We define the identity tensor as $\boldsymbol{\mathcal{I}}$ having ones on the hyper-diagonal and zero elsewhere. The trace of a matrix is denoted as $\text{tr}(\mathbf{X}^{I \times I}) = \sum_{i=1}^{I} x_{i,i}$. Furthermore, let $\mathbf{X}_{(n)} \in \mathbb{R}^{I_n \times I_1 \cdots I_{n-1} I_{n+1} \cdots I_N}$ be the n-mode matricization of $\boldsymbol{\mathcal{X}}$, $\text{vec}(\boldsymbol{\mathcal{X}}) \in \mathbb{R}^{I_1 \cdot I_2 \cdots I_N \times 1}$ be the vectorization of $\boldsymbol{\mathcal{X}}$ and $\otimes, \odot$ and $\circ$ denote the Kronecker, Khatri-Rao, and element-wise product, respectively. Let further $\times_n$ denote the n-mode matrix multiplication, such that $(\boldsymbol{\mathcal{X}} \times_n \boldsymbol{M})_{(n)} = \boldsymbol{M} \mathbf{X}_{(n)}$. For details on tensor notation and these operators see also [2].

### 2.1 Multi-way / Tensor Structure

Acquisition of data along multiple modes (time, samples, features, sensors, etc) results in multi-way/tensor data $\boldsymbol{\mathcal{X}}$ defined by a data hyper-cube. Here we concern our-self with multi-way data which specifically has an underlying multi-way/tensor structure, such as CPD, Tucker, or BTD or can be well approximated by such structure.

The Tucker model is given by,

$$\boldsymbol{\mathcal{X}} = \boldsymbol{\mathcal{G}} \times_1 \mathbf{A}^{(1)} \times_2 \mathbf{A}^{(2)} \times_3 \cdots \times_N \mathbf{A}^{(N)} + \boldsymbol{\mathcal{E}}, \tag{1}$$

$$x_{i_1, i_2, \ldots, i_N} = \sum_{d_1=1, d_2=1, \ldots, d_n=1}^{D_1, D_2, \ldots, D_N} g_{d_1, d_2, \ldots, d_N} a_{i_1, d_1}^{(1)} a_{i_2, d_2}^{(2)} \cdots a_{i_N, d_N}^{(N)} + \epsilon_{i_1, i_2, \ldots, i_N}, \tag{2}$$

where $i_n = 1, 2, \ldots, I_n$ are the observations in mode $n$ for $n = 1, 2, \ldots, N$, the factor matrices $\mathbf{A}^{(n)} \in \mathbb{R}^{I_n \times D_n}$ for mode $n$, the core array $\boldsymbol{\mathcal{G}} \in \mathbb{R}^{D_1 \times D_2 \times \cdots \times D_N}$ which defines the multi-linear ($N - linear$) interactions between the factors, and the residual error $\boldsymbol{\mathcal{E}}$ or $\epsilon_{i_1, i_2, \ldots, i_N}$. The Tucker model is highly expressive, but is not unique as it suffers from rotational ambiguity which makes interpretation of the determined structure difficult. In contrast, the CPD model is unique under mild conditions, thus having only scaling and permutation ambiguity [2]. Although, the CPD model is generally treated as a separate model, it can be viewed as a restricted Tucker model, specifically, requiring the same number of components in each mode, $D_n = D \forall_n$ and letting the core array be a hyper-unit array, e.g. $\boldsymbol{\mathcal{G}} = \boldsymbol{\mathcal{I}}$ where $i_{d,d,\ldots,d} = 1, d = 1, 2, \ldots, D$ and zero otherwise thereby only admitting interactions between components of same index. Thus the CPD can be written as

$$\boldsymbol{\mathcal{X}} = \boldsymbol{\mathcal{I}} \times_1 \mathbf{A}^{(1)} \times_2 \mathbf{A}^{(2)} \times_3 \cdots \times_N \mathbf{A}^{(N)} + \boldsymbol{\mathcal{E}}, \tag{3}$$

The Block-Term Decomposition (BTD) model was proposed in [5] and can be considered as a sum of Tucker models or as a special case of the Tucker model where the core array has a (multi-linear) block diagonal structure, i.e.,

$$\boldsymbol{\mathcal{X}}^{I_1 \times I_2 \times \cdots \times I_N} = \sum_t^T \boldsymbol{\mathcal{G}}_t \times_1 \mathbf{U}_t^{(1)} \times_2 \mathbf{U}_t^{(2)} \times_3 \cdots \times_N \mathbf{U}_t^{(N)} + \boldsymbol{\mathcal{E}}$$

$$= \tilde{\boldsymbol{\mathcal{G}}} \times_1 \tilde{\mathbf{U}}^{(1)} \times_2 \tilde{\mathbf{U}}^{(2)} \times_3 \cdots \times_N \tilde{\mathbf{U}}^{(N)} + \boldsymbol{\mathcal{E}},$$

where $\boldsymbol{\mathcal{G}}_t \in \mathbb{R}^{D_t^{(1)} \times D_t^{(2)} \times \cdots \times D_t^{(N)}}$ for each core $t = 1, 2, \ldots, T$ and orthogonal factor matrix $\mathbf{U}_t^{(n)} \in \mathbb{R}^{I_n \times D_t^{(n)}}$ for $t = 1, 2, \ldots, T$ and $n = 1, 2, \ldots, N$, and $\boldsymbol{\mathcal{E}}$ is the residual error. The multi-linear block diagonal formulation has $\tilde{\boldsymbol{\mathcal{G}}} = \texttt{blkdiag}\left(\boldsymbol{\mathcal{G}}_1, \boldsymbol{\mathcal{G}}_2, \ldots, \boldsymbol{\mathcal{G}}_T\right)$ and $\tilde{\mathbf{U}}^{(n)} = [\mathbf{U}_1^{(n)}, \mathbf{U}_2^{(n)}, \ldots, \mathbf{U}_T^{(n)}]$ for $n = 1, 2, \ldots, N$. Again, $I_n$ is the number of observations in the $n^{\text{th}}$-mode and $D_t^{(n)}$ is the number of latent components in the $n^{\text{th}}$-mode of the $t^{\text{th}}$-core. The $\texttt{blkdiag}$-operator take the $T$ cores and stacks them along the hyper-diagonal to yield a single core array with size $D_n = \sum_{t=1}^T D_t^{(n)}$.

The BTD is unique for 3-way data when sub-cores has size $(L_t, L_t, 1)$ where $L_t$ is the number of components for the $t^{\text{th}}$ core $\boldsymbol{\mathcal{G}}_t$ [5]. For BTD models without this special core structure, the uniqueness is not guaranteed and it is necessary to use either post-hoc core rotation schemes or apply regularization to improve interpretability of the core, similar to the Tucker models.

## 2.2 Bayesian Inference and Tensor Decomposition

Most tensor decomposition models has, historically, been devised under the least squares error and solved by least squares optimization, e.g. maximum likelihood estimation under noise following a Gaussian distribution. The maximum likelihood estimation (MLE) has been extended to other losses/distributions (binary, Poisson, etc.) which improves the modelling options, but does not solve the fundamental issues that MLE only provides a point estimate of the parameters $\boldsymbol{\theta}$. In contrast, parameter estimation using Bayesian inference goes beyond a point estimate and approximates the posterior distribution of the parameters given the data. The contrast between MLE and Bayesian methods is exemplified through Bayes rule,

$$P(\boldsymbol{\theta}|\boldsymbol{\mathcal{X}}) = P(\boldsymbol{\mathcal{X}}|\boldsymbol{\theta})P(\boldsymbol{\theta})P(\boldsymbol{\mathcal{X}})^{-1} \tag{4}$$

where $P(\boldsymbol{\mathcal{X}}) = \int_{\boldsymbol{\theta}} P(\boldsymbol{\mathcal{X}}|\boldsymbol{\theta})P(\boldsymbol{\theta})d\boldsymbol{\theta}$. Note, the left side is the posterior distribution and the right side is the likelihood, the prior, and the inverse of evidence or marginal distribution, respectively. Determining the parameters via MLE uses only the likelihood, $P(\boldsymbol{\mathcal{X}}|\boldsymbol{\theta})$, whereas Bayesian inference characterizes the distribution of the parameters and accounts for uncertainties in relation to the observed data and the prior information. For a detailed background on Bayesian inference, the reader is referred to [9].

Bayesian tensor decomposition - also known as probabilistic tensor decomposition - have advance both CPD and Tucker to a fully Bayesian treatment, thus allowing uncertainty quantification through the posterior distribution, a principled approach for incorporation of prior information and random variables, and automatic penalization of model complexity. These extensions have shown that Bayesian methods provide more robust parameter estimates, can handle a lower signal to noise ratio, are less sensitive to missing data, and the automatic penalization of model complexity defends against over-fitting [3]. Additional Bayesian extensions of tensor models include different likelihoods and different prior distributions on the factor matrices, as well as novel or reformulated tensor models, such as tensor train decomposition, PARAFAC2, multi-tensor factorization, collapsed Tucker, and many more. For an introduction to Bayesian tensor models, see [3, 4]

The key challenge in Bayesian inference is that calculating the posterior is generally intractable, as calculating the marginal distribution or evidence, $P(\boldsymbol{\mathcal{X}})$ , is intractable. Therefore, most applications of Bayesian inference turn to approximate methods, see [9], which are primarily; 1) Markov Chain Monte Carlo (MCMC) sampling and its variants. 2) Laplace approximation and its variants. 3) Variational approximation. We presently consider the variational approximation using variational Bayesian inference.

For variational Bayesian inference, the main idea is to approximate the posterior, $\log P(\boldsymbol{\theta}|\mathbf{X})$, using a set of variational distributions, $Q(\boldsymbol{\theta})$. The $Q$ distributions provide a lower-bound on the log evidence $\log P(\boldsymbol{\mathcal{X}})$ which is called the

evidence lower bound (ELBO). Using Jensen's inequality, the ELBO is,

$$\log P(\mathcal{X}) = \log \int_{\boldsymbol{\theta}} P(\mathcal{X}, \boldsymbol{\theta}) \frac{Q(\boldsymbol{\theta})}{Q(\boldsymbol{\theta})} d\boldsymbol{\theta}$$

$$\geq \int_{\boldsymbol{\theta}} Q(\boldsymbol{\theta}) \log \frac{P(\mathcal{X}, \boldsymbol{\theta})}{Q(\boldsymbol{\theta})} d\boldsymbol{\theta} = \texttt{ELBO}(\mathcal{Q}) \tag{5}$$

The variational distributions are chosen such that the ELBO is tractable [9]. We presently consider a mean-field approximation where the probability distributions are assumed to be independent between subsets of random variables, i.e. $Q(\boldsymbol{\theta}) = \prod_i Q(\boldsymbol{\theta}_i)$. The optimal variational distribution for each subset $\boldsymbol{\theta}_i$ is then,

$$Q(\boldsymbol{\theta}_i) \propto \exp\left\{ \langle \ln P(\mathcal{X}, \boldsymbol{\theta}) \rangle_{\setminus Q(\boldsymbol{\theta}_i)} \right\} \tag{6}$$

where $\langle \cdot \rangle_{\setminus Q(\boldsymbol{\theta}_i)}$ is the expected value with respect to all variational distributions except $Q(\boldsymbol{\theta}_i)$, see [9]. The optimal parameters of the variational distribution are then identified via moment matching in eq. (6).

We presently highlight Bayesian inference in the context of the Block Term Decomposition (BTD) which naturally encompass the Tucker model, i.e., $T = 1$, and CPD models, i.e., $T = D$ and $D_1^{(1)} = D_1^{(N)} = \ldots = D_T^{(1)} = D_T^{(N)} = 1$. The presented BTD is in itself a novel probabilistic tensor factorization approach explicitly imposing orthogonality on the factor loadings as conventional BTD and Tucker decompostions but also serves as illustration of probabilistic modeling in the context of tensors decompositions highlighting the steps from formulating a generative model, to inferring parameters accounting for uncertainty as opposed to conventional maximum likelihood estimation (MLE) for tensor decompositions relying on point estimates defined by $\boldsymbol{\theta}^* = \arg\max_{\boldsymbol{\theta}} P(\mathcal{X}|\boldsymbol{\theta})$.

## 2.3 Probabilistic Block Term Decomposition

The probabilistic or fully Bayesian Block Term Decomposition model was first proposed in [4], but it was not applied to any data, nor was its inferrential properties investigated. Both points will be remedied in this paper. Recently, an approach to Bayesian BTD was proposed by [8], but only considering $(L_r, L_r, 1)$ blocks and by formulating the factors as normal distributions, which is not inline with MLE BTD [5] imposing orthogonality and does not benefit from the inferrential properties afforded by all-orthogonal factors. We presently consider the following generative model for BTD,

$$P(\tau) = \mathcal{G}(\tau|\alpha_\tau, \beta_\tau), \tag{7}$$

$$P(\mathbf{U}_t^{(n)}) = \text{v}\mathcal{MF}(\mathbf{U}_t^{(n)}|\mathbf{F}_0), \; n = 1, \ldots, N, \; t = 1, \ldots, T, \tag{8}$$

$$P(\boldsymbol{\Psi}_t|\cdot) = \textbf{see eq. } (12), t = 1, \ldots, T, \tag{9}$$

$$P(\mathcal{G}_t) = \mathcal{N}\left(\texttt{vec}(\mathcal{G}_t)|\mathbf{0}, \boldsymbol{\Psi}_t^{\langle-1\rangle}\right) \tag{10}$$

$$P(\mathcal{X}|\boldsymbol{\theta}) = \mathcal{N}\left(\texttt{vec}(\mathcal{X})|\texttt{vec}\left(\sum_{t=1}^T \mathcal{M}_t\right), \tau^{-1}\mathbf{I}_J\right), \tag{11}$$

where $\mathcal{M}_t = \mathcal{G}_t \times_1 \mathbf{U}_t^{(1)} \times_2 \mathbf{U}_t^{(2)} \times_3 \cdots \times_N \mathbf{U}_t^{(N)}$ is the structure of the $t^{\text{th}}$ core, $\tau$ is the noise precision, and $J = \prod_{n=1}^N I_n$. The factor matrices are orthogonal and arises from the von Mises-Fisher matrix distribution, $\text{v}\mathcal{MF}(\mathbf{F}_0)$, where the a priori assumption is a uniform distribution on the Stiefel manifold, i.e. $\mathbf{F}_0 = \mathbf{0}$. The core arrays $\mathcal{G}_t$ are modelled by normal distributions with zero mean and precision matrix $\boldsymbol{\Psi}_t$ for sub-core $t$. For the noise precision, we let $\alpha_\tau = \beta_\tau = 1e - 3$ using the shape-rate formulation of the Gamma distribution.

The parameters are $\boldsymbol{\theta} = \{\tau, \{\mathbf{U}_t^{(n)}\}_{\forall t, \forall n}, \{\mathcal{G}_t, \boldsymbol{\Psi}_t\}_{\forall t}\}$ which we will call the noise precision, the factor matrices, the core arrays, and the penalization prior on the cores, respectively.

When specifying the prior on the core precision, we view the BTD model as a Tucker model, $\tilde{\mathcal{G}} = \texttt{blkdiag}(\mathcal{G}_1, \mathcal{G}_2, \ldots, \mathcal{G}_T)$, and vectorize the core array,

$$\texttt{vec}\left(\tilde{\mathcal{G}}\right) \sim \mathcal{N}\left(\texttt{vec}\left(\tilde{\mathcal{G}}\right)|\mathbf{0}, \boldsymbol{\Psi}^{-1}\right), \tag{12}$$

where $\boldsymbol{\Psi}^{K \times K}$ is the precision matrix with $K = \prod_{t=1}^T \prod_{n=1}^N D_t^{(n)}$. All the considered core priors have zero mean, as such different functionality is only obtained by changing the specification of $\boldsymbol{\Psi} = \texttt{blkdiag}(\boldsymbol{\Psi}_1, \ldots, \boldsymbol{\Psi}_T)$. Similar to pTucker in the probabilistic tensor toolbox [3], we can consider estimating scale, element-wise sparsity, or pruning irrelevant slices of each core which can be formulated as,

$$\boldsymbol{\Psi} = \texttt{blkdiag}(\psi_1 \mathbf{I}_{\mathbf{D}_1}, \psi_2 \mathbf{I}_{\mathbf{D}_2}, \ldots, \psi_T \mathbf{I}_{\mathbf{D}_T}), \tag{13}$$

$$\boldsymbol{\Psi} = \texttt{blkdiag}(\text{diag}(\boldsymbol{\psi}_1), \text{diag}(\boldsymbol{\psi}_2), \ldots, \text{diag}(\boldsymbol{\psi}_T)), \tag{14}$$

$$\boldsymbol{\Psi} = \texttt{blkdiag}(\bigotimes_{n=1} \text{diag}(\boldsymbol{\psi}_1^{(n)}), \ldots, \bigotimes_{n=1} \text{diag}(\boldsymbol{\psi}_T^{(n)})), \tag{15}$$

where $\mathbf{D}_t = \prod_{n=1}^{N} D_t^{(n)}$ and each precision $\psi_t$ follows a shape-rate gamma distribution with $\alpha_\psi = \beta_\psi = 1e-3$. The sparsity prior has a precision on each element of the BTD cores, e.g. $\boldsymbol{\psi}_t = [\psi_{t,1}, \psi_{t,2}, \ldots, \psi_{t,\mathbf{D}_t}]$, and the ARD prior has precision $\boldsymbol{\psi}_t^{(n)} = [\psi_{t,1}^{(n)}, \psi_{t,2}^{(n)}, \ldots, \psi_{t,D_t}^{(n)}]$ thus $D_t^{(n)}$ elements for each core $t$. For a detailed discussion on the different prior types see [3, 4].

Our implementation supports the scale and sparsity prior, but for space concerns we presently only consider the sparsity prior. For the variational inference we impose the following factorized Q-distribution

$$Q(\boldsymbol{\theta}) = Q(\tau) \prod_t P(\boldsymbol{\mathcal{G}}_t) Q(\boldsymbol{\Psi}_t) \prod_n Q(\mathbf{U}_t^{(n)}) \tag{16}$$

### 2.3.1 Determining the factor matrices

Deriving the update rule for the factor matrix belonging to sub-core $t$ and mode $n$, we observe that the Q-distribution $Q(\mathbf{U}_t^{(n)})$ also follows a v$\mathcal{MF}$. The resulting concentration matrix $\tilde{\mathbf{F}}_t^{(n)}$ is given by

$$\tilde{\mathbf{F}}_t^{(n)} = \langle \tau \rangle \; \langle \mathbf{G}_{t,(n)} \rangle \left( \bigotimes_{m \neq n} \langle \mathbf{U}_t^{(m)} \rangle \right) \tilde{\mathbf{Y}}^\top + \mathbf{F}_0, \tag{17}$$

$$\tilde{\mathbf{Y}} = \mathbf{X}_{(n)} - \sum_{t' \neq t} \langle \mathbf{G}_{t',(n)} \rangle \left( \bigotimes_{m \neq n} \langle \mathbf{U}_{t'}^{(m)} \rangle \right), \tag{18}$$

where $\tilde{\mathbf{Y}}$ is the observed data but with the influence of the $t' \neq t$ cores subtracted, $\mathbf{G}_{t,(n)}$ is the $t$-core matricized along mode $n$. The optimal variational distribution according to eq.(6) is thus $Q(\mathbf{U}_t^{(m)}) = \text{v}\mathcal{MF}\left( \mathbf{U}_t^{(m)} | \tilde{\mathbf{F}}_t^{(n)} \right)$.

### 2.3.2 Determining the core or sub-cores

The core arrays are updated one by one, here formulated as the vectorized sub-core $\texttt{vec}(\boldsymbol{\mathcal{G}}_t) \equiv \mathbf{g}_t$ which comes a priori from a normal distribution and the Q-distribution also follows a normal distribution. Thus, the update is,

$$\boldsymbol{\Sigma}_{\mathbf{g}_t} = \left( \langle \boldsymbol{\Psi}_t \rangle + \langle \tau \rangle \; \left\langle \left( \bigotimes_{n=1}^{N} \mathbf{U}_t^{(n)^\top} \mathbf{U}_t^{(n)} \right) \right\rangle \right)^{-1}, \tag{19}$$

$$\boldsymbol{\mu}_{\mathbf{g}_t} = \boldsymbol{\Sigma}_{\mathbf{g}_t} \; \langle \tau \rangle \left( \bigotimes_{n=1}^{N} \langle \mathbf{U}_t^{(n)} \rangle \right)^\top \mathbf{y}, \tag{20}$$

where $\mathbf{y} = \texttt{vec}(\boldsymbol{\mathcal{X}}) - \sum_{t' \neq t} \left( \bigotimes_{n=1}^{N} \langle \mathbf{U}_{t'}^{(n)} \rangle \right) \langle \mathbf{g}_{t'} \rangle$ is the data without the parts modelled by the $t' \neq t$ blocks, $\texttt{vec}(\boldsymbol{\mathcal{G}}_t) \equiv \mathbf{g}_t \in \mathbb{R}^{D_t^{(1)} D_t^{(2)} \cdots D_t^{(N)} \times 1}$ and $\boldsymbol{\Psi}_t$ the prior on its precision matrix. The variational distribution is $Q(\mathbf{g}_t) = \mathcal{N}(\mathbf{g}_t | \boldsymbol{\mu}_{\mathbf{g}_t}, \boldsymbol{\Sigma}_{\mathbf{g}_t})$. Notably, the orthogonal factor matrices $\mathbf{U}_t^{(n)^\top} \mathbf{U}_t^{(n)} = \mathbf{I}$ allow efficient calculation of the core covariance $\boldsymbol{\Sigma}_{\mathbf{g}_t}$ as $\bigotimes_{m \neq n} \mathbf{U}_t^{(m)^\top} \mathbf{U}_t^{(m)} = \mathbf{I}_{K_t}$ where $K_t = \prod_{m \neq n} D_t^{(m)}$ resulting in simple univariate distributions of each core element, i.e. the elements of the $t^{th}$ core are independent such that $Q(\mathbf{g}_t) = \prod_{c=1}^{D_t^{(1)} D_t^{(2)} \cdots D_t^{(N)}} Q(\mathbf{g}_t(c))$.

### 2.3.3 Determining the model penalization

Since each of the core priors can be represented as a diagonal matrix, we will let $g_t \sim \mathcal{N}(\mathbf{0}, \text{diag}(\boldsymbol{\psi}_t)^{-1})$ and show how the scale, sparsity, and automatic relevance determination (ARD) prior are updated. Since the elements of $\boldsymbol{\psi}_t$ come from an i.i.d. shape-rate Gamma distribution $\mathcal{G}(\alpha_\psi, \beta_\psi)$ and the residual error is normal distributed, then the Q-distribution also follows a Gamma distribution $Q(\psi_*) \sim \mathcal{G}(\tilde{\alpha}_{\psi_*}, \tilde{\beta}_{\psi_*})$,

$$\tilde{\alpha}_\psi = \alpha_\psi + 0.5 \prod_{n=1}^N D_n \tag{21}$$

$$\tilde{\alpha}_{\psi_d} = \alpha_\psi + 0.5 \tag{22}$$

$$\tilde{\alpha}_{\psi_{n,d}} = \alpha_\psi + 0.5 \prod_{n' \neq n}^N D_n \tag{23}$$

where $\psi_*$ consists of a single element, $\sum_{t=1}^T \prod_{n=1}^N D_t^{(n)}$ elements, and $\sum_{t=1}^T \sum_{n=1}^N D_t^{(n)}$ elements for the scale, sparsity, and ARD prior, respectively. The constant $0.5$ comes from the specification of $P(\psi_*)$ as a Gamma distribution, which can easily be changed to an exponential prior giving a constant of $1$, but having the same Q distribution, see [3].

The rate parameter depends on the value of the core array and changes between iterations, it can be formulated as

$$\tilde{\beta}_{\psi_t} = \beta_\psi + 0.5 \left\langle \mathbf{g}_t^\top \mathbf{g}_t \right\rangle, \tag{24}$$

$$\tilde{\beta}_{\psi_{t,\mathbf{d}}} = \beta_\psi + 0.5 \left\langle g_{t,\mathbf{d}} \right\rangle \quad \mathbf{d} = [d_1^t, d_2^t, \ldots, d_N^t], \tag{25}$$

$$\tilde{\beta}_{\psi_{t,d_n^t}^{(n)}} = \beta_\psi + 0.5 \mathrm{Tr} \left( \left\langle G_{t,(n),d_n^t}^\top G_{t,(n),d_n^t} \right\rangle \bigotimes_{n' \neq n} \mathrm{diag}(\left\langle \boldsymbol{\psi}_t^{(n)} \right\rangle) \right), \tag{26}$$

where $\mathbf{d}$ indexes a single element in $\mathbf{g}_t$ and $G_{t,(n),d_n^t}$ is a vector containing all $\mathbf{g}_t$ core elements which interact with the parameter $\psi_{t,d_n^t}^{(n)}$.

### 2.3.4 Determining the noise precision

The noise distribution follows a Gamma prior and similar for the Q-distribution $Q(\tau) \sim \mathcal{G}(\tilde{\alpha}_\tau, \tilde{\beta}_\tau)$,

$$\tilde{\alpha}_\tau = \alpha_\tau + \frac{I_1 \cdots I_N}{2} \tag{27}$$

$$\tilde{\beta}_\tau = \beta_\tau + \frac{1}{2} \left( \mathbf{x}^\top \mathbf{x} + \left\langle \tilde{\mathbf{g}}^\top \mathbf{I} \tilde{\mathbf{g}} \right\rangle + \mathbf{x}^\top \left( \bigotimes_{n=1}^N \left\langle \mathbf{U}_t^{(n)} \right\rangle \right) \left\langle \tilde{\mathbf{g}} \right\rangle \right) \tag{28}$$

where $\mathbf{x} = \mathrm{vec}(\boldsymbol{\mathcal{X}})$ and $\tilde{\mathbf{g}} = \mathrm{vec}(\tilde{\boldsymbol{\mathcal{G}}})$ and $\mathbf{I}$ is the identity matrix due to the orthogonal factors. Presently, we only consider homoscedastic noise and refer to [3] for heteroscedastic noise in probabilistic tensor decomposition.

## 3 Results

We first investigate using either Bayesian and maximum likelihood estimation (MLE) based BTD to learn the BTD structure of synthetic generated data. Next, we illustrate the use of the probabilistic BTD (pBTD) on synthetic data with known ground truth and on two real-world dataset to evaluate whether a CPD, Tucker, or a BTD model defining intermediates is the best structure. We will use the formulation BTD($C$,$D$) where $C$ is the number of cores and $D$ is the size of each subcore, e.g. $\mathcal{G}_c \in \mathbb{R}^{D \times D \times D}$, $c = 1, 2, \ldots, C$. For each study, we fit six BTD models with varying number of equally sized cores, specifically we fit $(C, D) = (12, 1), (6, 2), (4, 3), (3, 4), (2, 6)$, and $(1, 12)$. The first model corresponds to a CPD model without orthogonal factors (as orthogonality is within each single component core) and the last model to a Tucker model with twelve orthogonal components.

### 3.1 Synthetic Study - MLE vs Bayesian BTD

Data was generated according to a $\boldsymbol{\mathcal{M}}_{true} = BTD(4, 3)$ model and homoscedastic Gaussian noise was added to vary the signal-to-noise ratio from $-20db$ to $30dB$ in steps of $2.5dB$. The resulting data $\boldsymbol{\mathcal{X}} = \boldsymbol{\mathcal{M}}_{true} + \boldsymbol{\mathcal{E}}$ was then decomposed to obtain $\boldsymbol{\mathcal{M}}_{est}$ with the correct model order (BTD($4, 3$)) and with an incorrect model order BTD($4, 6$) for the proposed probabilistic BTD (pBTD) and the two MLE BTD methods from the Tensorlab Toolbox [10] denoted BTD-NLS and BTD-minf. The methods were compared by their relative reconstruction error on the noiseless data, e.g. $||\mathrm{vec}(\boldsymbol{\mathcal{M}}_{true}) - \mathrm{vec}(\boldsymbol{\mathcal{M}}_{est})|| / ||\mathrm{vec}(\boldsymbol{\mathcal{M}}_{true})||_2$ where $|| \cdot ||_2$ is the Euclidean norm.

The experiment was repeated 100 times to asses the variability of the results, each time with new true data and noise generated. The results, Figure 1a, show that for the correct model order all methods have similar performance. Here it is worth noting, that for high noise levels, pBTD learns to turn off the modelling - whereas the MLE approaches overfit noise in the data and has high loss on the noiseless data. For the incorrect model order - though still knowing the correct number of blocks - the pBTD model is preferred as it achieves a lower error. The pBTD model has higher variability between runs than BTD-minf and BTD-NLS. The BTD-minf model is the fastest model with pBTD being 1.5x to 2x times slower and BTD-NLS is by far the slowest being 20-30x slower than BTD-minf. The proposed pBTD method achieves better results than both MLE methods and is 10-15x times faster than BTD-NLS (the best performing MLE method).

We next assess if the pBTD model can be used to determine which BTD structure a dataset follows. We simulate a ground truth model,. $\mathcal{M}_{true} = \text{BTD}(C, D)$ with additive Gaussian noise at SNR$= 10dB$ for each of six BTD models, e.g. $(C, D) = (12, 1), (6, 2), (4, 3), (3, 4), (2, 6)$, and fit the same six BTD models to each generated model. The relative reconstruction error against the noiseless data and the evidence lowerbound (ELBO, eq. (5)) are calculated.

The ELBO and relative reconstruction error for the best performing models (highest ELBO) out of 100 repeated fittings are shown in Figure 1b. Overall, the pBTD model is able to identify the correct BTD structure via ELBO and this structure also achieves the best reconstruction error. This is not always the case, as seen for pBTD-$(4, 3)$ where the models with fewer but larger cores obtain higher ELBO and lower error. Generally, we see the pruning ability based on the sparse score prior (eq. (14)) is able to get a good performance out of an over-specified model.

We investigate the case with a ground truth BTD-$(4, 3)$ model further and show the estimated pBTD models in Figure 1 bottom panel. Here, the core arrays of the BTD $(4, 3), (3, 4)$, an $(2, 6)$ are especially interesting, as the show many elements are pruned to near zero thus mitigating the issue of model misspecification.

### 3.2 Real Data

We apply the pBTD model on two real datasets and investigate the model range pursued in the synthetic studies, e.g. from a CPD to Tucker model with four intermediate steps. To avoid local minima, the model fitting is repeated ten times and the model obtaining the highest ELBO is displayed. The data sets considered are the Flow Injection dataset described in [11] and the electroencephalogram (EEG) dataset described in [12].

The Flow Injection dataset [11] comes from a industrial injection molding process and consist of 12 samples, 100 wavelengths, and 89 timepoints, e.g. $\mathcal{X} \in \mathbb{R}^{12 \times 100 \times 89}$. The data is freely available at www.ucphchemometrics.com/datasets. The results are shown in Figure 2 top panel. Notably, inspecting the ELBO we find support for the Tucker model but also observe that the inferred Tucker core has many elements close to zero which we attribute to the pruning of the automatic relevance determination imposed on each core element. Notably, some of the components of the Tucker models also have more uncertainty associated to the component than the more constrained BTD and in particular CPD representations when inspecting the factor loadings norms (not shown).

The electroencephalogram (EEG) dataset of inter-trial phase coherence (ITPC) [12] consists of EEG data from a 64-channel electrodearray acquired for 14 subject during stimulation of left and right hand (28 measurements). For each measurement, the signal from each channel and trial is measured in the time and frequency domain using a continuous wavelet transform and the ITPC calculated as the absolute value of the average phase across epochs which was then vectorized to a 4392 signal. This gives rise to a $\mathcal{X}^{28 \times 64 \times 4392}$ tensor, that we center across the channels mode before the analysis. The results are shown in Figure 2 bottom panel. From the ELBO we again observe highest support for the full Tucker model indicating a preference for a full multilinear representation of the data and we also here observe although less pronounced pruning of core elements by the automatic relevance determination. Contrary to the Tucker and BTD representations the CPD provides more easily interpretable representations with notably the first and eleventh component well discriminating between left and right hand stimulation.

## 4 Conclusion

We presented the probabilistic Block Term Decomposition (pBTD) and highlighted how variational Bayesian inference can be used in the context of tensor factorization using the BTD also naturally encompassing the two perhaps most prominent tensor factorization models CPD and Tucker as special cases.

On synthetic data we highlighted the utility accounting for uncertainty and noise providing added robustness to overfitting when compared to conventional maximum likelihood estimation. We further found that the probabilistic BTD as opposed to convention BTD had added robustness to model mis-specification whereas the inferential complexity is similar to conventional MLE. Notably, the Bayesian inference further admits automatic relevance determination en-

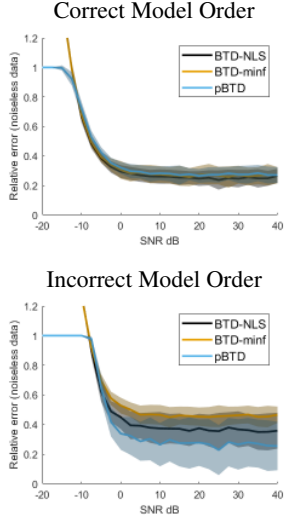

Correct Model Order

Incorrect Model Order

(a) Comparison of pBTD, BTD-NLS, and BTD-minf on reconstruction of the noiseless data at different Signal-to-Noise Ratios (SNR) in dB. For the correct BTD$(4, 3)$ and incorrect BTD$(4, 6)$ model order. The solid line is the mean and the shaded area is the standard deviation over 100 instances.

**Estimated pBTD-$(Cores, Components)$ model**

| | (12,1) | (6,2) | (4,3) | (3,4) | (2,6) | (1,12) |
|---|---|---|---|---|---|---|
| | Scaled ELBO (higher is better) | | | | | |
| BTD-(12,1) | **1.0000** | 0.9988 | 0.9969 | 0.9941 | 0.9861 | 0.6808 |
| BTD-(6,2) | 0.9952 | **1.0000** | 0.9984 | 0.9949 | 0.9860 | 0.9402 |
| BTD-(4,3) | 0.9414 | 0.9412 | 0.9858 | **1.0000** | 0.9963 | 0.9389 |
| BTD-(3,4) | 0.8327 | 0.8350 | 0.8782 | **1.0000** | 0.9444 | 0.9267 |
| BTD-(2,6) | 0.7014 | 0.7061 | 0.7311 | 0.7661 | **1.0000** | 0.9267 |
| BTD-(1,12) | 0.4298 | 0.4320 | 0.4456 | 0.4636 | 0.5087 | **1.0000** |
| | Relative Reconstruction Error (lower is better) | | | | | |
| BTD-(12,1) | **0.0453** | **0.0453** | 0.0458 | 0.0468 | 0.0498 | 1.0000 |
| BTD-(6,2) | 0.0683 | **0.0478** | 0.0484 | 0.0488 | 0.0516 | 0.0660 |
| BTD-(4,3) | 0.2248 | 0.2244 | 0.1334 | 0.0766 | **0.0513** | 0.0653 |
| BTD-(3,4) | 0.3808 | 0.3807 | 0.3023 | **0.0483** | 0.1683 | 0.0658 |
| BTD-(2,6) | 0.5389 | 0.5332 | 0.4890 | 0.4302 | **0.0497** | 0.0647 |
| BTD-(1,12) | 0.7355 | 0.7314 | 0.7113 | 0.6827 | 0.6103 | **0.0681** |

(b) Simulated (row) vs estimated (column) BTD structure. Shown is the scaled ELBO (1 is best) and relative error (0 is best) of the best performing (highest ELBO) pBTD models. The pBTD model is able to assess the correct structure in most cases, except for BTD-$(4, 3)$.

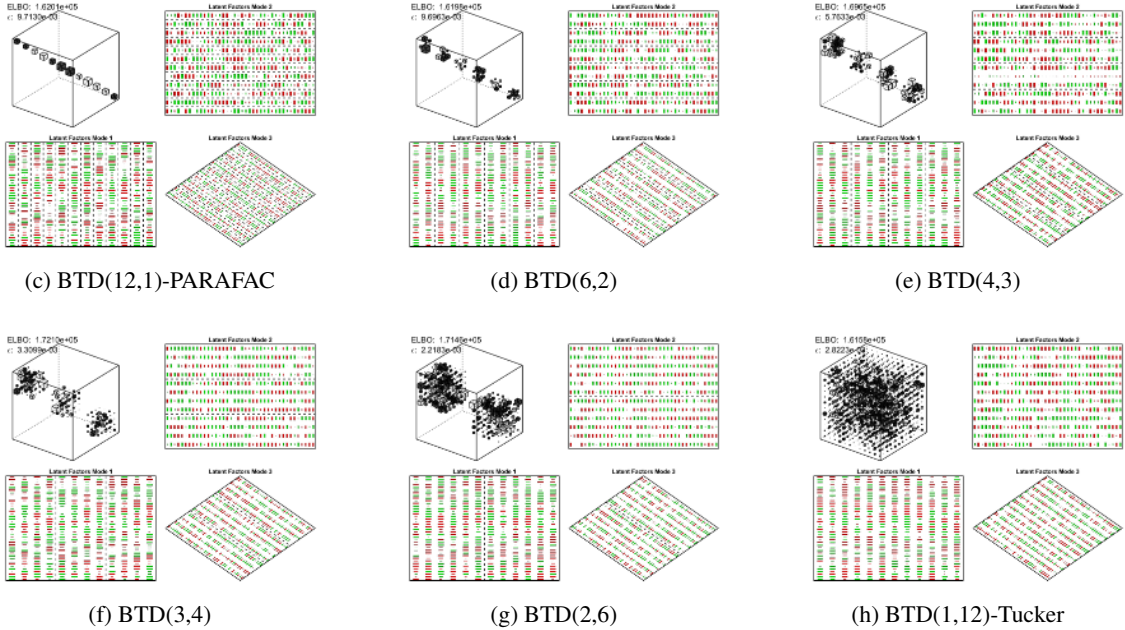

(c) BTD(12,1)-PARAFAC

(d) BTD(6,2)

(e) BTD(4,3)

(f) BTD(3,4)

(g) BTD(2,6)

(h) BTD(1,12)-Tucker

Figure 1: **Synthetic studies**: (a) Compares Bayesian vs MLE based BTD at multiple SNR levels. (b) Determining the underlying BTD structure of a dataset using pBTD. (c-h) The estimated pBTD models when the true structure is BTD$(4, 3)$. Each decomposition is visualized as a Tucker model with a core array and three factor matrices with mode 1, 2, and 3 visualized as the bottom left, top right, and bottom right plot. The ELBO and relative error $\epsilon$ are superimposed in the top right corner.

**Flow Injection Data**

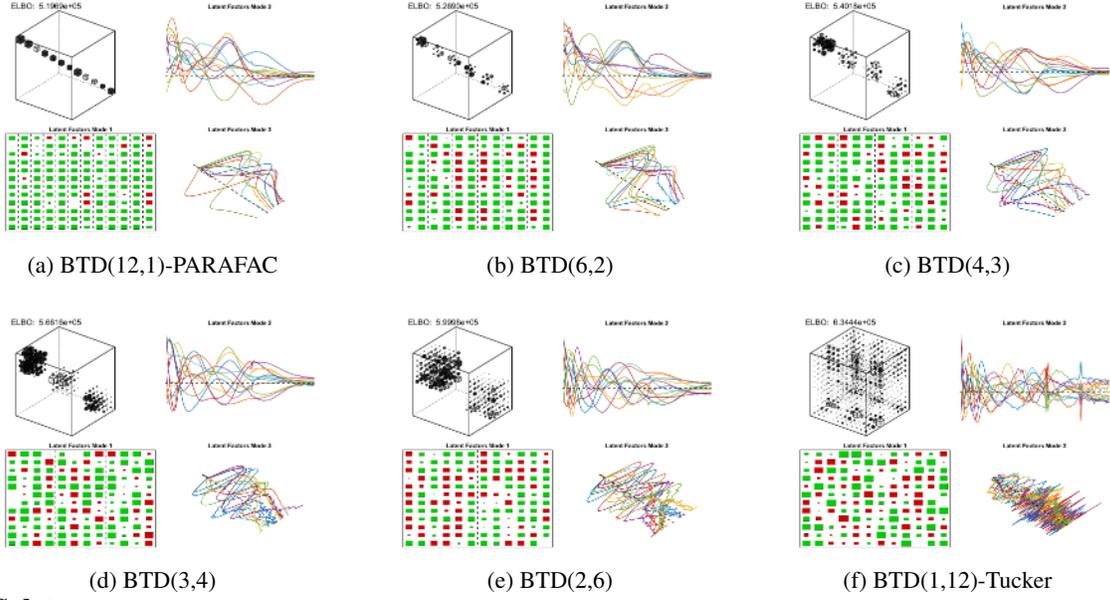

(a) BTD(12,1)-PARAFAC     (b) BTD(6,2)     (c) BTD(4,3)

(d) BTD(3,4)     (e) BTD(2,6)     (f) BTD(1,12)-Tucker

**EEG data**

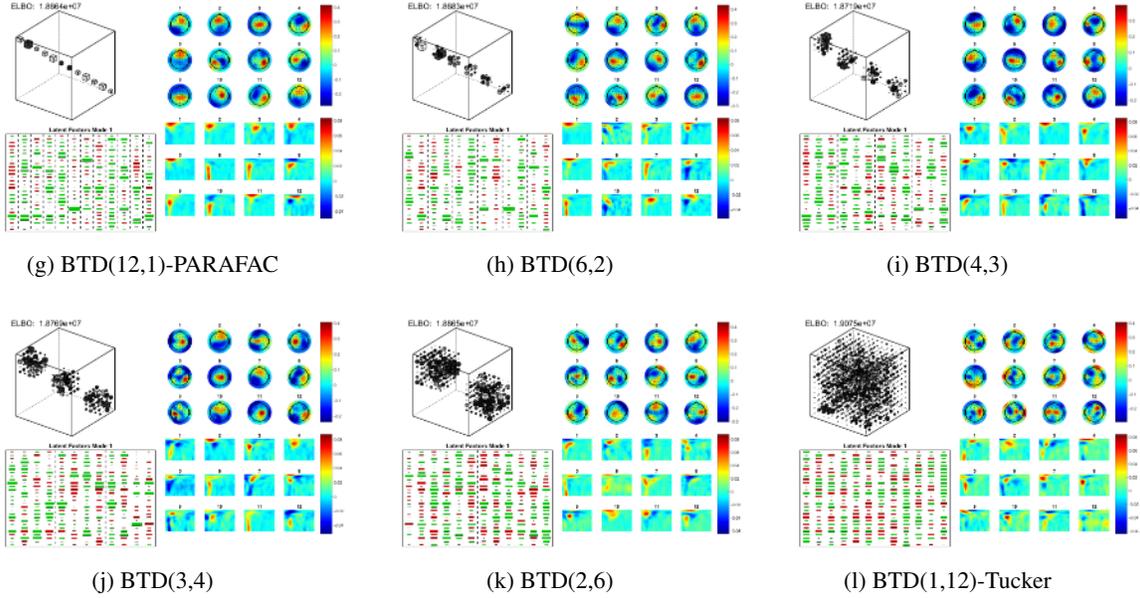

(g) BTD(12,1)-PARAFAC     (h) BTD(6,2)     (i) BTD(4,3)

(j) BTD(3,4)     (k) BTD(2,6)     (l) BTD(1,12)-Tucker

Figure 2: **Top panel:** Flow injection data $\mathcal{X} \in \mathbb{R}^{12 \times 100 \times 89}$ decomposed by six different pBTD models. Each decomposition is visualized as a Tucker model with a core array and three factor matrices with mode 1, 2, and 3 visualized as the bottom left, top right, and bottom right plot. The black dashed lines in mode 2 and 3 indicate zero on the y-axis. The ELBO is superimposed in the top right corner. **Bottom panel:** Wavelet transformed inter-trial phase coherence (ITPC) EEG data $\mathcal{X} \in \mathbb{R}^{28 \times 64 \times 4392}$ decomposed by six different pBTD models. Each decomposition is visualized as a Tucker model with a core array and three factor matrices with mode 1, 2, and 3 visualized as the bottom left, top right, and bottom right plot. For mode 2 and 3, the components are shown individually as topographic plots (scalp maps) and image of the refolded time-frequency representation, respectively. The ELBO is superimposed in the top right corner.

abling the pruning of core elements by learning their length scales as well as model assessment through the associated evidence lower bound (ELBO) and we found that this enabled successful identification of model structures imposed in the synthetically generated data.

On real data we further highlighted how the pBTD provided sound representations enabling explicit noise modeling and component quantification accounting for parameter uncertainty. We further observed how automatic relevance determination imposed on each elements of the core separately appeared to reduce many of the core components to become small and further found that the ELBO favored the full Tucker representations for the two considered data sets. However, from an interpretation point of view the full Tucker model representation as well as the intermediate BTD representation are harder to interpret when compared to the CPD due to component interactions within blocks. As such, we observe for the EEG data that component one and eleven for the CP model well discriminates left stimulation from right stimulation, whereas the other model representations did not provide such simple interpretable representations.

In summary, we provided a unified probabilistic framework for CPD, Tucker and BTD and demonstrated merits of Bayesian inference in the context of tensor factorization providing added robustness to model misspecification and overfitting with efficient inference similar in complexity to conventional MLE. Notably, such probabilistic inference admits component pruning of core elements using automatic relevance determination and model assessment through the associated ELBO as well as explicit noise modeling and uncertainty quantification of model parameters. We presently considered variational inference but we note that the framework readily generalizes to MCMC inference using Gibbs sampling by sampling from the posterior distribution as opposed to presently defining moments of the posterior used for the variational inference.

Importantly, the presented pBTD exemplifies the utility of Bayesian inference in the context of tensor decomposition highlighting how conventional tensor decomposition approaches, i.e., the BTD as presently considered, can be expanded to probabilistic modeling explicitly accounting for orthogonality with computational complexity similar to conventional BTD estimation. The proposed pBTD thereby contributes to recent efforts expanding tensor decomposition models to probabilistic inference.

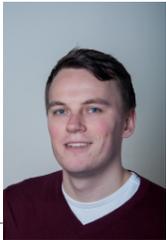

[ ]Jesper Løve Hinrich Jesper Løve Hinrich received the M.Sc. (2016) and Ph.D. (2020) degrees in applied mathematics from the Technical University of Denmark. He is currently a Postdoc at the Department for Food Science, University of Copenhagen, Denmark. His research interests includes multi-way modelling, Bayesian inference, statistical machine learning, and their application in the Life Sciences.

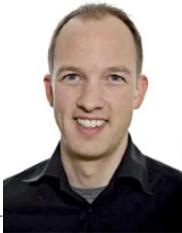

[ ]Morten Mørup received the MS and PhD degrees in applied mathematics from the Technical University of Denmark, Denmark, where he is currently professor at the Section for Cognitive Systems at DTU Compute. He has been associate editor of the IEEE Transactions on Signal Processing and his research interests include machine learning, neuroimaging, and complex network modeling.